\documentclass[11pt,a4paper]{article}
\usepackage[hyperref]{conll-2019}
\usepackage{times}
\usepackage{latexsym}
\usepackage{hyperref}
\usepackage{bm,amsmath,amssymb}
\usepackage{url}
\usepackage{booktabs}
\usepackage{graphicx}
\usepackage{makecell}
\usepackage{multirow}
\usepackage{todonotes}
\usepackage{subcaption}

\aclfinalcopy % Uncomment this line for the final submission

\newcommand{\Note}[2]{} 
\newcommand{\SideNote}[2]{} 
\renewcommand{\Note}[2]{\todo[color=#1,size=\small, inline=true]{#2}} 
\renewcommand{\SideNote}[2]{\todo[color=#1,size=\small]{#2}} % 

\title{Learning A Unified Named Entity Tagger From Multiple Partially Annotated Corpora For Efficient Adaptation}

\author{Xiao Huang$^{1,2}$, Li Dong$^3$\thanks{\quad Work done while the author was at USC ISI.}, Elizabeth Boschee$^{1}$, Nanyun Peng$^{1,2}$ \\
  $^1$Information Sciences Institute, University of Southern California \\
  $^2$Department of Computer Science, University of Southern California \\
  $^3$Outreach, Inc. \\
  {\tt \{huan183, lidong\}@usc.edu, \{boschee, npeng\}@isi.edu} \\}

\date{}

\begin{document}
\maketitle
\begin{abstract}
Named entity recognition (NER) identifies typed entity mentions in raw text.
While the task is well-established, there is no universally used tagset: often, datasets are annotated for use in downstream applications and accordingly only cover a small set of entity types relevant to a particular task.
For instance, in the biomedical domain, one corpus might annotate genes, another chemicals, and another diseases---despite the texts in each corpus containing references to all three types of entities.
In this paper, we propose a deep structured model to integrate these ``partially annotated'' datasets to {\em jointly} identify all entity types appearing in the training corpora.
By leveraging multiple datasets, the model can learn robust input representations; by building a joint structured model, it avoids potential conflicts caused by combining several models' predictions at test time.
Experiments show that the proposed model significantly outperforms strong multi-task learning baselines when training on multiple, partially annotated datasets and testing on datasets that contain tags from more than one of the training corpora.\footnote{The code and the datasets will be made available at \url{https://github.com/xhuang28/NewBioNer}}
\end{abstract}

\begin{figure*}[t!]
\centering
\includegraphics[width=\linewidth]{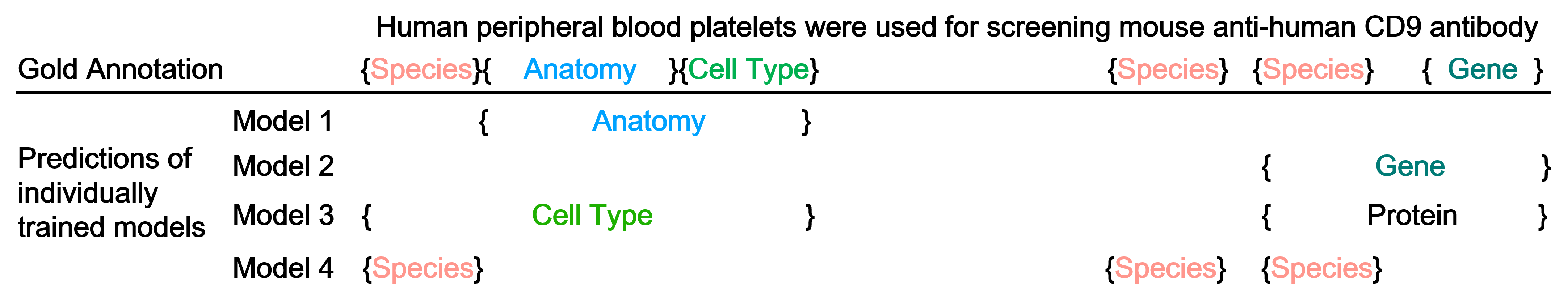}
\caption{An example sentence from the CellFinder corpus \cite{neves2012annotating} showing the challenges in combining the output of individual NE taggers. The \textit{Gold} row is the human annotations in CellFinder. The rows below are predictions made by models trained on datasets that each contain only a subset of the CellFinder types. Note that the individual taggers' predictions can conflict with each other, making it challenging to combine them. 
(Note: we renamed CellFinder's \textit{Cell Component} to \textit{Cell Type} to fit it in the space above.) }\label{fig:collision}
\vspace{-.5em}
\end{figure*}

\section{Introduction}
Named Entity Recognition (NER), which identifies the boundaries and types of entity mentions from raw text, is a fundamental problem in natural language processing (NLP).
It is a basic component for many downstream tasks, such as relation extraction \cite{hasegawa2004discovering,mooney2005subsequence}, coreference resolution \cite{soon2001machine}, and knowledge base construction \cite{craven1998learning,craven1999constructing}. 

One problem in NER is the diversity of entity types, which vary in scope for different domains and downstream tasks. 
Traditional NER for the news domain focuses on three coarse-grained entity types: \textit{person}, \textit{location}, and \textit{organization}~\cite{tjong2003introduction}. However, as NLP technologies have been applied to a broader set of domains, many other entity types have been targeted. For instance, \newcite{ritter2011named} add seven new entity types (e.g., \textit{product}, \textit{tv-show}) on top of the previous three when annotating tweets. Other efforts also define different but partially overlapping sets of entity types \cite{
%grishman1995design,
walker2006ace,ji2010overview,ere13,aguilar2014comparison}. 
These non-unified annotation schemas result in \emph{partially annotated datasets}: each dataset is only annotated with a subset of possible entity types. 

One approach to this problem is to train individual NE taggers for each partially annotated dataset and combine their results using some heuristics. Figure~\ref{fig:collision} shows an example that demonstrates the possible shortcomings of this approach, using the biomedical domain as a case study.\footnote{\url{https://corposaurus.github.io/corpora/} summarizes dozens of partially annotated biomedical datasets.}
Here, we train four separate models on four partially annotated datasets:  AnatEM~\cite{pyysalo2013anatomical} annotated for the \emph{anatomy} type, BC2GM~\cite{smith2008overview} for the \emph{gene} type, JNLPBA~\cite{kim2004introduction} for \emph{cell} types, and Linnaeus~\cite{gerner2010linnaeus} for the \emph{species} type. 
We can see that the models' predictions contradict each other when applied to the same test sentence---making it a challenge to accurately combine them.

%In this paper, we propose to build a single, unified model from multiple partially  annotated datasets to \emph{jointly} identify all entity types that appear in any of the training sets.  The model leverages supervision signals across diverse datasets to learn robust feature representations, thus improving the performance for each entity type. Moreover, it makes {\em joint} predictions to avoid potential conflicts among models built on different entity types, allowing further improvement in the cross-type NER results. 

%The contributions of the paper are threefold:
%\vspace{-0.5em}
%\begin{itemize}
%\itemsep-.2em 
%\item We propose a deep structured model to leverage multiple partially annotated datasets, building a unified model that jointly identifies the union of all entity types represented in the training data.
%\item We design a novel ``global evaluation'' to gauge the effectiveness of the unified model, even when no single unified test set exists.
%\item Compared to strong multi-task learning baselines, experiments on the biomedical domain show that our model achieves comparable results when evaluated on individual datasets (i.e.\ identifying proteins in a corpus annotated for proteins), and significantly better results when evaluated on datasets with more types than any individual dataset (i.e\ identifying both proteins and genes in a test set using a model trained on a protein-only dataset and a gene-only dataset).
%\end{itemize}

In this paper, we propose a deep structured model to leverage multiple partially annotated datasets, allowing us to jointly identify the union of all entity types presented in the training data. The model leverages supervision signals across diverse datasets to learn robust input representations, thus improving the performance for each entity type. Moreover, it makes {\em joint} predictions to avoid potential conflicts among models built on different entity types, allowing further improvement for cross-type NER. 

Experiments on both real-world and synthetic datasets show that our model can efficiently adapt to new corpora that have more types than any individual dataset used for training and that it achieves significantly better results compared to strong multi-task learning baselines.
%evaluated on the datasets that have more types than any individual dataset used for training. 

\section{Problem Statement}

We formally define the problem by first defining our terminology. 
%Let $T_U$ denote the set of all possible entity types in the universe.

\textbf{Global Tag Space.} Let $C_i$ denote a corpus, and $T_{C_i}$ denote the set of entity types that are tagged in corpus $C_i$. When there are a set of corpora $\mathbb{C} = \{C_{1}, C_{2}, ... , C_{n}\}$, each has its own tag space concerning different entity types, the global tag space is defined as the union of the local tag space. Formally, $T_{\mathbb{C}} = T_{C_1} \cup T_{C_2} \cup ... \cup T_{C_n}$. 

\textbf{Partially Annotated Corpus.} If $T_{C_i} \subsetneqq T_{\mathbb{C}}$, then $C_i$ is a partially annotated corpus.

%\textbf{Local Evaluation.} When a model is trained on a partially annotated corpus $C_{i}$, and makes predictions within $T_{C_i}$, we say it making {\em local predictions}. Accordingly, when the model is only evaluated on $T_{C_i}$, it is called {\em local evaluation}.

\textbf{Global Evaluation.} When a model is trained on a set of partially annotated corpora $\mathbb{C}$ and predicts tags for the whole global tag space $T_{\mathbb{C}}$, we say it is making {\em global predictions}. Accordingly, the evaluation of the models' performance on $T_{\mathbb{C}}$ is called {\em global evaluation}.

Our goal is to train a single unified NE tagger from several \textit{partially annotated} corpora for efficient adaptation to new  corpora  that  have  more  types  than  any individual dataset used  during training. 
Formally, we have a set of corpora $\mathbb{C} = \{C_{1}, C_{2}, ... , C_{n}\}$, and we propose to train a joint model on $\mathbb{C}$ such that it makes predictions for the \emph{global tag space} $T_{\mathbb{C}}$.
One benefit of this joint model is that it can be easily adapted to a new tag space $T_{C_u}$ where $T_{C_u} \subseteq T_{\mathbb{C}}$, and $T_{C_u} \nsubseteq T_{C_i}, \forall C_i \in \mathbb{C}$. %|T_{C_u}| > |T_{C_i}|
%\NoteNP{I changed this and I think this is more broad}

\section{Background and Related Work}
\label{sec:background}
In this section, we first introduce neural architectures for NER which our work builds upon and then summarize previous work on imperfect annotation problems. 

\subsection{Neural Architectures for NER}

With recent advances using deep neural networks, bi-directional long short-term memory networks with conditional random fields (BiLSTM-CRF) have become standard for NER \cite{lample2016neural}. A typical architecture consists of a BiLSTM layer to learn feature representations from the input and a CRF layer to model the inter-dependencies between adjacent labels and perform joint inference. 
\newcite{ma2016end} introduce additional character-level convolutional neural networks (CNNs) to capture subword unit information. 
In this paper, we use a BiLSTM-CRF with character-level modeling as our base model. 
We now briefly review the BiLSTM-CRF model.

\paragraph{BiLSTMs.} Long Short Term Memory networks (LSTMs)~\cite{hochreiter1997long} are a variation of RNNs that are designed to avoid the vanishing/exploding gradient problem~\cite{bengio1994learning}. 
Specifically, BiLSTMs take as input a sequence of words
%$\bm{x}=\{x_1, x_2,..., x_n\}$ 
$\bm{x}=\{x_k|k \in \mathcal{N}\}$ 
and output a sequence of hidden vectors: 
%$H = \{\bm{h}_1, \bm{h}_2,..., \bm{h}_n\}$. 
$\bm{H}=\{h_k|k \in \mathcal{N}\}$ 
BiLSTMs combine a left-to-right (forward) and a right-to-left (backward) LSTM to capture both left and right context. Formally, they produce a hidden vector $\tilde{\bm{h}}_i = [\overrightarrow{\bm{h}_i}; \overleftarrow{\bm{h}_i}]$ for each input, where $\overrightarrow{\bm{h}_i}$ and $\overleftarrow{\bm{h}_i}$ are produced by the forward and the backward LSTMs respectively; $[;]$ denotes vector concatenation.

\paragraph{Character-level Modeling.}
Following \newcite{wang2018cross}, we use a BiLSTM
%to model character-level information. 
for character-level modeling. 
We concatenate the hidden vector of the space after a word from the forward LSTM and the hidden vector of the space before a word from the backward LSTM to form a character-level representation of the word: $\bm{h}^c_i = [\overrightarrow{\bm{h}_i^c};\overleftarrow{\bm{h}^c_i}]$.
The word-level BiLSTM then takes the concatenation of $\bm{h}^c_i$ and the word embedding as input $\bm{x}_i = [\bm{e}_i;\bm{h}^c_i]$ to learn contextualized representations.

\paragraph{Neural-CRFs.}
Conditional Random Fields (CRFs) \cite{lafferty2001conditional} are sequence tagging models that capture the inter-dependencies between the output tags; they have been widely used for NER \cite{mccallum2003early,lu2015chemdner,pengnamed,pengimproving,peng2016multi}. 
Given a set of training data $\{\bm{x}_i, \bm{y}_i\}^N$, a CRF minimizes negative log-likelihood:

%\small{
\begin{gather}
\min_\Theta - \sum_{i} \log P(\bm{y}_i\mid \bm{x}_i; \Theta), \label{eq:orig-lh} \\
%\end{equation}
%\begin{equation}\label{eq:orig-lh-rewrite}
%\small
P(\bm{y}_i\mid \bm{x}_i; \Theta) = \frac{Gold \, Energy}{Partition} = \frac{St(\bm{y}_i)}{\sum_{\bm{y}'} St(\bm{y}')} \label{eq:orig-lh-rewrite}
\end{gather} %}
\normalsize
where $\bm{y}'$ is any possible tag sequence with the same length as $\bm{y}_i$, $St(\bm{y}')$ is the potential of the tag sequence $\bm{y}'$, and $St(\bm{y}_i)$ is the potential of the gold tag sequence. The numerator $St(\bm{y}_i)$ is called the \emph{gold energy function}, and the denominator $\sum_{\bm{y}'} St(\bm{y}')$ is the \emph{partition function}. The likelihood function using globally annotated data is illustrated in Figure~\ref{fig:orig-lh}.
The potential of a tag sequence can be computed as:
\begin{equation}\label{eq:orig-pt}
%\small
St(\bm{y}) = \prod_{t=1}^{|\bm{y}|} Score(\bm{y}[t],\bm{y}[t-1])
\end{equation}%~\NoteNP{changed t to start from 1.}
where $\bm{y}[t]$ is the $t$th element in $\bm{y}$ ($\bm{y}[-1]$ is the start of the sequence), and
\begin{equation}\label{eq:orig-score}
%\small
\begin{split}
Score(\bm{y}[t],\bm{y}[t-1]) = & \exp\left(tr(\bm{y}[t],\bm{y}[t-1])\right) * \\
 & \exp\left(em(\bm{y}[t])\right)
\end{split}
\end{equation}
where $tr(\bm{y}[t],\bm{y}[t-1])$ is the transition score from $\bm{y}[t-1]$ to $\bm{y}[t]$, and $em(\bm{y}[t])$ is the emission score of $\bm{y}[t]$ computed based on the output $\tilde{\bm{h}}_t$ of the BiLSTM.

\subsection{Learning from Imperfect Annotations}
Learning from multiple partially annotated datasets could be more generally thought of as learning from imperfect annotations. In that broad sense, there are several notable areas of prior work. One of the most prominent concerns learning from \textit{incomplete annotations} (noisy labels), where some occurrences of entities are neglected in the annotation process and falsely labeled as non-entities (negative). A related problem is learning from \textit{unlabeled data} with distant supervision. 

A major challenge of all these settings, including ours, is that a positive instance might be labeled as negative. A well-explored solution to this problem is proposed by \newcite{C08-1113}, which %is a special case of our general framework and will be explained in Section~\ref{sec:improve-energy}. In their proposal,
instead of maximizing the likelihood of the gold tag sequence,  we maximize the total likelihood for all possible tag sequences consistent with the gold labels. \newcite{C08-1113,Yang2014SemiSupervisedCW} applied this idea to the \emph{incomplete annotation} setting; \newcite{shang2018learning,D14-1093} applied it to the \emph{unlabeled data} with distant supervision setting; and \newcite{greenberg2018marginal} applied it to the \emph{partial annotation} setting. 
While this is a general solution, its primary drawback is that it assumes a uniform prior on all labels consistent with the gold labels. This may have the result of overly encouraging the prediction of entities, resulting in low precision.

To tackle the problem of incomplete annotations,  \newcite{Carlson2009LearningAN,yang2018distantly} explored bootstrap-based semi-supervised learning on unlabeled data, iteratively identifying new entities with the taggers and then re-training the taggers. \newcite{bellare2007learning,li2005learning,fernandes2011learning} explored an EM algorithm with semi-supervision.

For the \emph{partial annotation} problem,
most previous work has focused on building individual taggers for each dataset and using single-task learning \cite{liu2017empower} or multi-task learning \cite{crichton2017neural,wang2018cross}. In single-task learning, each model is trained separately on each dataset $C_{i}$, and makes local predictions on $T_{C_i}$. Based on the neural-CRF architecture, multi-task learning uses a different CRF layer for each dataset $C_{i}$ (each task) to make local predictions on $T_{C_i}$, and shares the lower-level representation learning component across all tasks. Both single-task learning and multi-task learning make local predictions and have to apply heuristics to combine the model predictions, resulting in the collision problem demonstrated in Figure~\ref{fig:collision}. 

To the best of our knowledge, \newcite{greenberg2018marginal} is the only prior work trying to build a unified model from multiple partially annotated corpora. We will show that their model, which is reminiscent of \newcite{C08-1113}, is a special case of ours and that our other variations achieve better performance. In addition, they only evaluated the model on the training corpora while we conduct evaluations to test the model's ability to adapt to new corpora with different tag spaces.

\section{Model}

As mentioned above, we use a BiLSTM-CRF with character-level modeling as our base model. Our goal is to build a unified model to make global predictions. That is, our model will be jointly trained on multiple partially annotated datasets $\mathbb{C}$ and make predictions on the global tag space $T_{\mathbb{C}}$. 
Such a unified model will enjoy the benefit of learning robust representations from multiple datasets just like multi-task learning while maintaining a joint probability distribution of the global tag space to avoid possible conflicts from individual models.

\subsection{Naive Approach} \label{sec:naive}
\begin{figure}[t!]
    \centering
    \begin{subfigure}[t]{0.48\textwidth}
        \includegraphics[width=\textwidth]{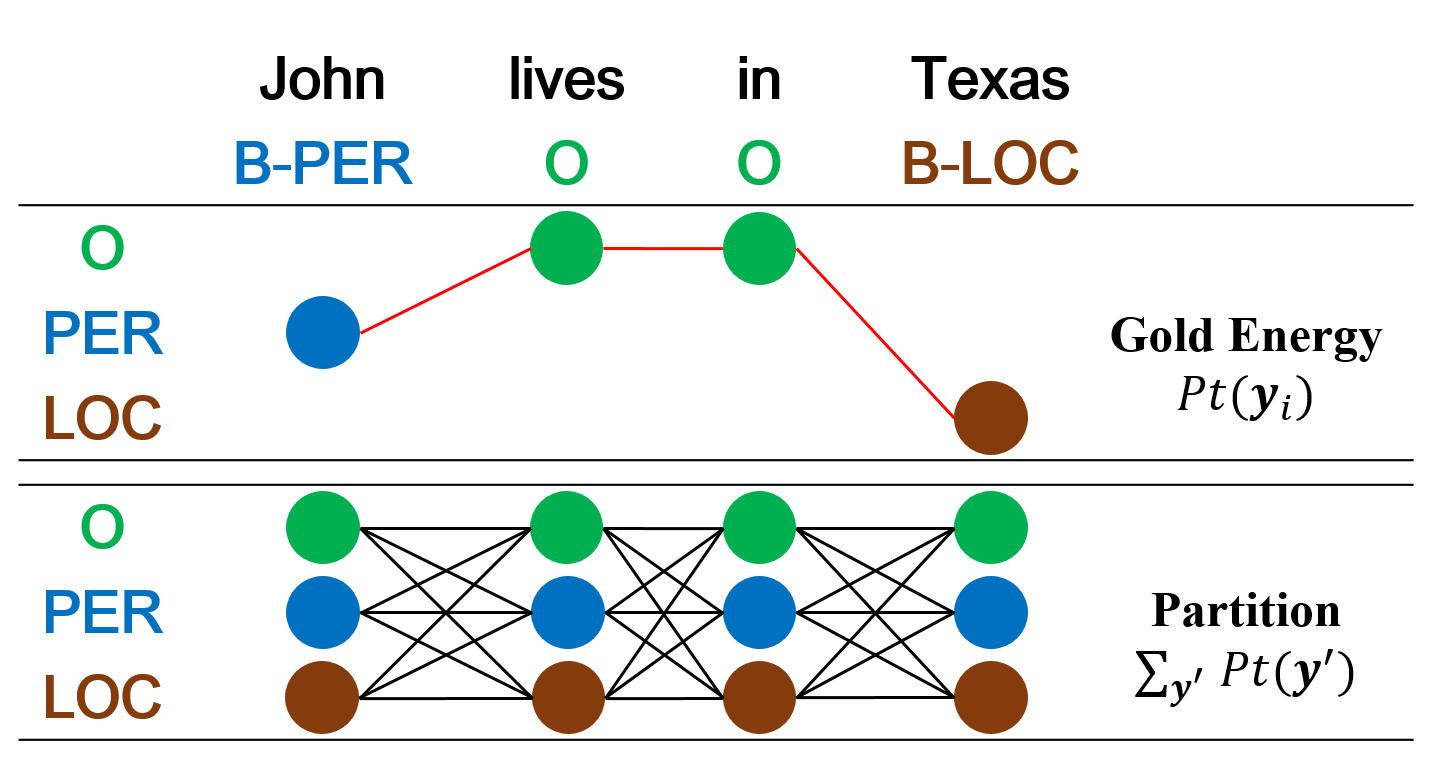}
        \caption{\textbf{Original likelihood function} with \textit{global annotation}}
        \label{fig:orig-lh}
    \end{subfigure}
    \\
    \begin{subfigure}[t]{0.48\textwidth}
        \includegraphics[width=\textwidth]{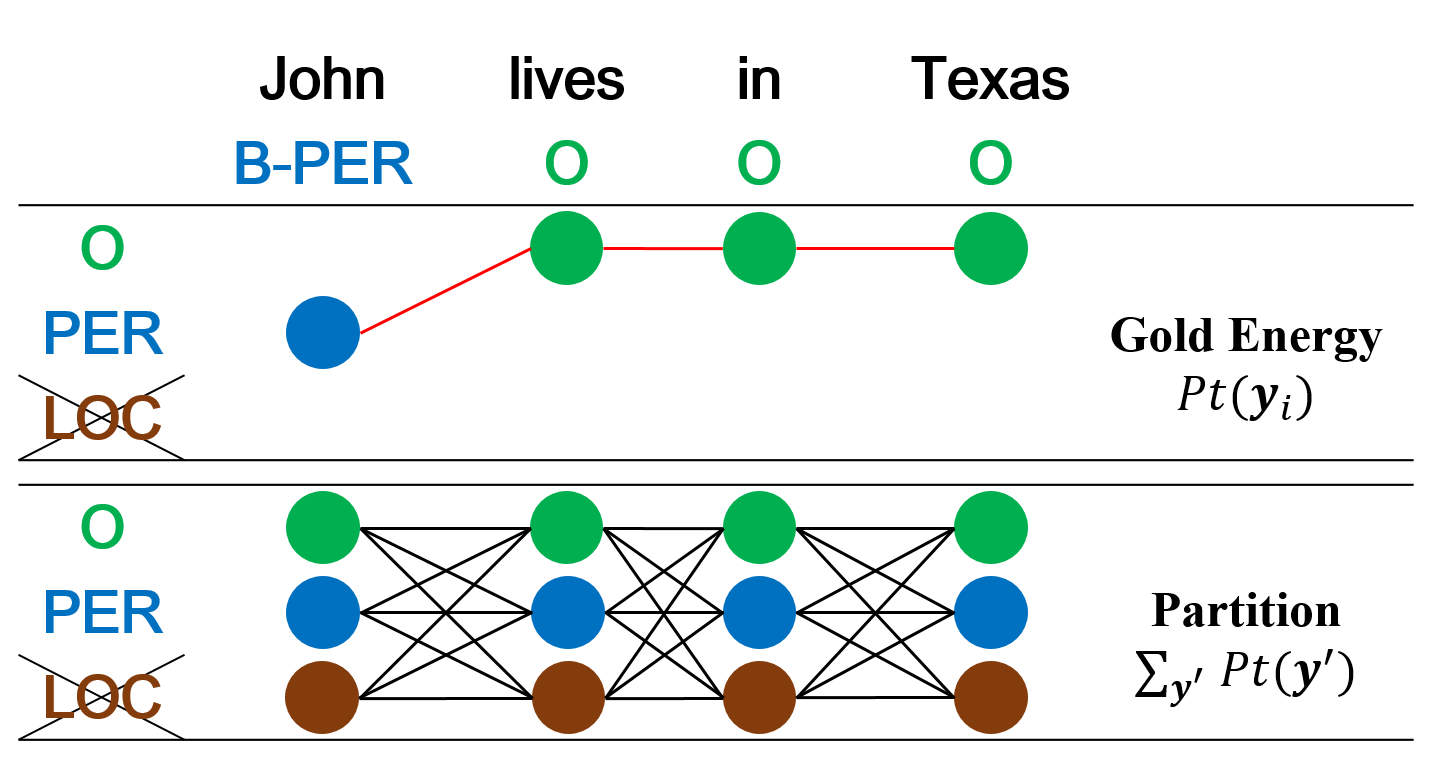}
        \caption{\textbf{Original likelihood function} with \textit{partial annotation}}
        \label{fig:naive-lh}
    \end{subfigure}
    \\
    \begin{subfigure}[t]{0.48\textwidth}
        \includegraphics[width=\textwidth]{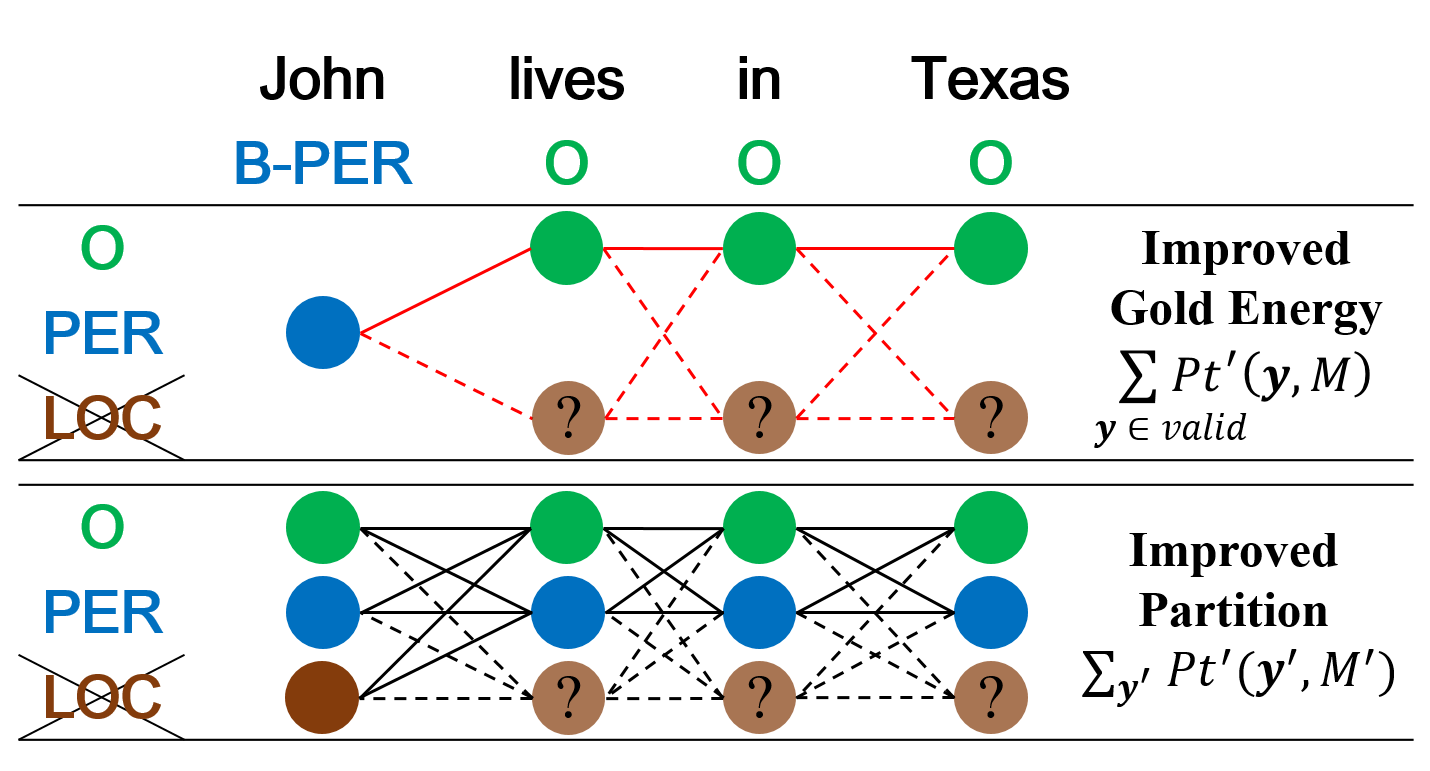}
        \caption{\textbf{Improved likelihood function} with \textit{partial annotation}}
        \label{fig:improved-lh}
    \end{subfigure}
    \caption{Illustration of original~(\ref{fig:orig-lh},~\ref{fig:naive-lh}) and improved~(\ref{fig:improved-lh}) likelihood functions. 
    Each figure has two parts upper and lower that illustrate the gold energy (numerator) and the partition (denominator) respectively. Solid lines represent tag sequences that are fully considered in the functions. Dashed lines represent tag sequences that are discounted. The sentences in \ref{fig:naive-lh} and \ref{fig:improved-lh} are not annotated with \textbf{LOC}.}\label{fig:model}
    \vspace{-1em}
\end{figure}

A simple solution to the problem is to merge all the datasets into one giant corpus. 
A single model can then be trained on this corpus to make global predictions. However, such a corpus will be missing many correct annotations, since each portion will be annotated with only a subset of the target entity types. 
Figure~\ref{fig:naive-lh} shows an example: here, a location (\textit{Texas}) exists but is labeled as a non-entity, because the original dataset from which this sentence is drawn does not annotate locations at all. As a result, this approach suffers from {\em false penalties} when applying the original likelihood function (Eq. \ref{eq:orig-lh-rewrite}-\ref{eq:orig-score}) to train the model, meaning that it penalizes predictions that correctly identify entities with types that are not annotated for a particular sentence. 

\subsection{Improving the Gold Energy Function} \label{sec:improve-energy}
One way to improve performance is to explicitly acknowledge the incompleteness of the existing ``gold'' annotations and to give the model credit for predicting any tag sequence that is consistent with the partial annotations. 
This can be done by modifying the CRF's gold energy function, illustrated in the upper part of Figure~\ref{fig:improved-lh}. Specifically, in this example, \textit{John} is labeled as {\em PER}, so {\em PER} is the only possible correct tag at that position. However, \textit{lives}, \textit{in}, and \textit{Texas} are labeled as {\em O} (non-entity), which here means only that they may not be {\em PER}---but any of them could be {\em LOC}, since locations are not annotated for this sentence. Therefore, any sequence that assigns either {\em O} or {\em LOC} for any of these three positions is consistent with the gold labels. 
To account for this, we modify the gold energy function to credit all tag sequences that are consistent with the gold annotations, encouraging the model to predict other consistent labels when the gold label is {\em O}. 
\newcite{C08-1113} propose a specific solution that applies this idea on incomplete annotations: instead of maximizing the likelihood of the gold tag sequence when optimizing the CRF model, they maximize the total likelihood of all possible tag sequences consistent with the gold labels. This approach is later used by \newcite{greenberg2018marginal} to handle the problem of partial annotation. We will address a potential problem with their method and propose a generalized version in Section~\ref{sec:discAlter}.

\subsection{Improving the Partition Function} \label{sec:improve-partition}
Modifying the gold energy function will give credit to a system for producing alternative entity labels for words tagged as {\em O} in the partially annotated training. A different solution is to simply \textit{not penalize} predictions of such alternative labels.
This can be done by modifying the partition function and keeping the gold energy function unchanged. The lower part of Figure~\ref{fig:improved-lh} gives an illustration. As stated above, {\em LOC} is a consistent alternative entity label for \textit{lives}, \textit{in}, and \textit{Texas}. We therefore exclude from our calculations any paths that include {\em LOC} at any of those positions. More generally, we exclude all such consistent but alternative tag sequences from the computation of the CRF's partition function. Section~\ref{sec:discAlter} gives formal definitions with equations. 
The improved partition function sets the model free to predict alternative labels without penalty (as long as they are consistent with the known gold annotations), but it does not give them any positive credit for doing so (as in the previous approach).
We hypothesize that the improved partition function would work better than the improved gold energy function in our setting because it addresses the {\em false penalties} problem more precisely. We will verify this hypothesis in our experiments. 

\subsection{Discounting Alternative Sequences}
\label{sec:discAlter}
There is a potential problem with naively applying the improved gold energy function: when the gold label is {\em O}, the model is encouraged to predict other consistent labels as strongly as it is encouraged to predict {\em O}. However, many {\em O} labels are confident annotations of {\em O}. As a result, naively training with the improved gold energy function tends to over-predict entities and not predict {\em O}s.
To mitigate this problem, we discount the energy of tag sequences that go through alternative labels. This can be achieved by introducing a hyper-parameter $M \text{(mask)} \in [0,1]$ as a discounting factor for the gold energy function. Formally, we modify Eq~\ref{eq:orig-pt} to:

\vspace{-2em}
%{\small
\begin{align*}
& St'(\bm{y}, M) = \\ 
& \prod_{t=1}^{|\bm{y}|} (Score(\bm{y}[t],\bm{y}[t-1])* mask(\bm{y}[t], M)), 
\end{align*}
where 
$$ mask(\bm{y}[t], M) = 
    \begin{cases}
        %M, & \text{if } \bm{y}[t] \neq \hat{\bm{y}}[t] \\
        M, & \text{if } \bm{y}[t] \in alternative \\
        1, & \text{Otherwise}
    \end{cases}. $$
%}
where $alternative$ is the set of alternative labels.
We thus have the improved gold energy function:

%{\small
\begin{equation}
Improved \, Gold \, Energy = \sum_{\bm{y} \in valid} St'(\bm{y}, M),
\end{equation}
%}
where $valid$ is the set of all tag sequences that are consistent with the gold sequence, including the gold sequence itself.

Similarly, for the improved partition function, we can use the same strategy to discount the energy of alternative sequences rather than completely removing them. We thus introduce another $M' \in [0,1]$ and the improved partition function becomes:

%{\small
\begin{equation}
Improved \, Partition = \sum_{\bm{y}'} St'(\bm{y}', M'),
\end{equation}
%}
\subsection{Combining Improved Functions} \label{sec:combine}
For generality, we combine the improved gold energy and the improved partition function to make a new likelihood function as our final model:

\begin{equation}\label{eq:improved-lh}
%\small
Improved \, LH = \frac{\sum_{\bm{y} \in valid} St'(\bm{y}, M)}{\sum_{\bm{y}'} St'(\bm{y}', M')}
\end{equation}
To ensure Equation~\ref{eq:improved-lh} is a valid likelihood function (the probabilities of all sequences sum to 1), we need a constraint that $M = M'$. Note that Equation~\ref{eq:improved-lh} subsumes all models discussed in this section.
Specifically, when $M=0,M'=1$, the model is the {\em Naive Model} discussed in Section~\ref{sec:naive}; when $M=1,M'=1$, the model is the same as~\newcite{greenberg2018marginal} discussed in Section~\ref{sec:improve-energy}; when $M=0,M'=0$, the model is the same as proposed in Section~\ref{sec:improve-partition}. 
We have a general perspective of all the models by simply treating $M$ and $M'$ as hyper-parameters. 

Note that for the {\em Naive Model}, since $M != M$, the Equation~\ref{eq:improved-lh} is not always a valid likelihood function\footnote{This may be confusing because when $M=0, M'=1$ it looks exactly the same as the original CRF likelihood function. But in the partial annotation setting, this means that the scores of alternative sequences will be zero in the numerator but non-zero in the denominator, which makes the total likelihood less than 1. It suggests that the original CRF likelihood function is not suitable for the partial annotation setting.}. This may partially explain why the {\em Naive Model} performs so poorly under this setting. We posit that the model will work the best when $M=M'$. 

\section{Experimental Setup}
\subsection{Datasets}
Our goal is to train a unified NER model on multiple partially annotated datasets. This model will make global predictions and can efficiently adapt to new corpora that contain tags from more than one training corpus. To fully test this capability, we would need a single test set annotated with all types of interest. However, the motivation behind this effort is that such a dataset typically does not exist. We therefore take two approaches to approximate such an evaluation.

In the first evaluation setting, we take advantage of the fact that although there may not be a \emph{single} dataset annotated with all  types of named entities of interest, there exist several datasets that cover types from more than one of the training corpora. Specifically, we are able to select test corpora that each cover types of interest from multiple training corpora. Table~\ref{tab:data-ne} shows the biomedical corpora we use and their entity types. For example, we use \textbf{BC5CDR} for global evaluation, because its entity types (\textit{Chemical} and \textit{Disease}) cover multiple training corpora (\textbf{BC4CHEM} for \textit{Chemical} and \textbf{NCBI} for \textit{Disease}).
%Briefly, our five training corpora tagsets are effectively A, B, BC, C, ADEFG, and H. Our three test corpora tagsets are AEFH, AB, and ABC. 

\begin{table}[t]
\centering
%\small
\resizebox{\linewidth}{!}{
\begin{tabular}{|l | l|} 
\toprule
  \multicolumn{2}{|c|}{\textbf{For Training}} \\ \hline
  \textbf{Corpus} & \textbf{Entities} \\
  \hline
  BC2GM & GP \\ \hline
  BC4CHEM~~~~ & Chemical\\ \hline
  NCBI & Disease \\ \hline
  JNLPBA & \makecell[l]{GP,\\ DNA, \\Cell-type, \\Cell-line, \\RNA} \\ \hline
  Linnaeus & Species \\ 
  \bottomrule
\end{tabular}
\begin{tabular}{|l | l|}
  \toprule
  \multicolumn{2}{|c|}{\textbf{For Global Evaluations}} \\ \hline
  \textbf{Corpus} & \textbf{Entities} \\ 
  \hline
  BC5CDR & \makecell[l]{Chemical, \\Disease} \\ \hline
  BioNLP13CG & \makecell[l]{GP,  \\Disease, \\Chemical, \\others} \\  \hline
  BioNLP11ID & \makecell[l]{GP, \\Chemical, \\others} \\ 
  \bottomrule
\end{tabular}}

\caption{Details of the biomedical corpora. 
%``GP'' denotes Gene or Protein;
%, which are commonly used interchangeably; 
``others'' denotes NE types that do not appear in the training corpora, and thus are not evaluated.}
\label{tab:data-ne}
\end{table}

In the second evaluation setting, we create synthetic datasets from the CoNLL 2003 NER dataset to simulate training and global evaluations. Specifically, the CoNLL 2003 dataset is annotated with four entity types: location, person, organization, and miscellaneous entities. We randomly split the training set into four portions, each containing only one entity type (all other types are removed). In this setting, the four portions of the training set are used for training and the original dataset with all entities annotated is used as a global corpus.
%We argue that this setting better tests the models' ability to generalize to global tag space. 

More details about all the datasets can be found in Appendix~\ref{app:data}.

\begin{figure}[t!]
\centering
\includegraphics[width=.95\linewidth]{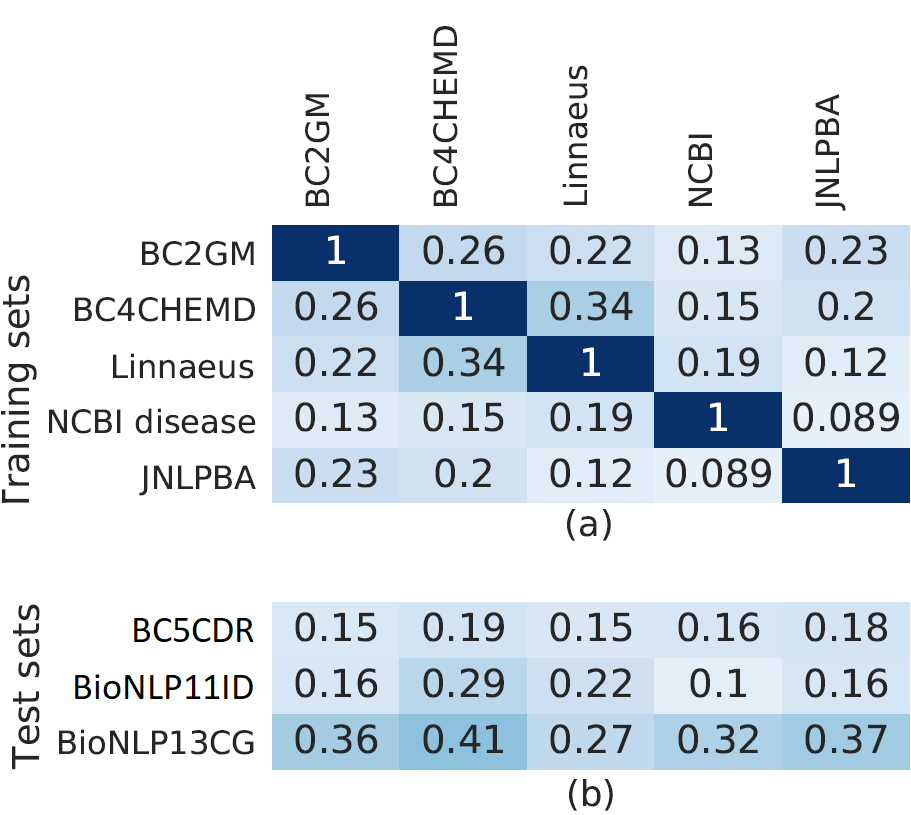}
\caption{(a) The mention-level overlap among training sets. (b) The mention-level overlap between training datasets and evaluation datasets.}\label{fig:overlap}
\vspace{-1em}
\end{figure}

\subsubsection{Biomedical Dataset Analysis}

The motivation for this work rests on the assumption that even when a dataset is annotated for a certain set of entity types, it likely contains other types of entities that are unlabeled.
To verify this assumption, we expand the annotations of each dataset using heuristics and compute the pairwise mention-level overlap between the datasets. 
Specifically, suppose we are comparing two datasets, A and B. We first construct A' and B', where A' contains all mentions in A but is augmented with new mentions found by taking all strings annotated in B and marking them as named entities in A (regardless of context; there may obviously be some errors). We do the same (in the opposite direction) to construct B'.
We then compute the pairwise overlap coefficient between A' and B' according to the following criterion: 
$$\mathrm{overlap}(A',B')={\frac {|A'\cap B'|}{\min(|A'|,|B'|)}}.$$ 
Figure~\ref{fig:overlap} shows the heat maps. For the training group, {\em BC2GM}, {\em BC4CHEMD}, and {\em Linnaeus} are considerably overlapped, although they are annotated with different entity types (GP, Chemical, and Species). 
This confirms our assumption that although the datasets are annotated for a subset of entity types, they contain other types that are unlabeled.\footnote{We further verified this conclusion by computing the heat maps on the original datasets. The overlaps between {\em BC2GM} and {\em BC4CHEMD}, and {\em BC2GM} and {\em Linnaeus} are nearly $0$.}

\begin{table*}[t!]
\centering
%\resizebox{\linewidth}{!}{
\begin{tabular}{l|*{3}{*{3}{l}|}*{3}{l}}%{l@{\ \ }|l@{\ \ } l@{\ \ } l@{\ \ }|l@{\ \ } l@{\ \ } l@{\ \ }|l@{\ \ } l@{\ \ } l@{\ \ }|l@{\ \ } l@{\ \ } l@{\ \ }}
\toprule
  \multirow{3}{*}{\textbf{Corpus}}& \multicolumn{9}{c|}{\bf Trained on Other Biomedical Datasets} & \multicolumn{3}{c}{\bf Traind on CoNLL} \\ \cline{2-13}
  %\cline(2-13}
   & \multicolumn{3}{c|}{\bf BC5CDR} & \multicolumn{3}{c|}{\bf BioNLP13CG} & \multicolumn{3}{c|}{\bf BioNLP11ID} &
  \multicolumn{3}{c}{\bf CoNLL 2003}\\
  & \multicolumn{1}{c}{F\ \ } & \multicolumn{1}{c}{P\ \ } & \multicolumn{1}{c|}{R} & \multicolumn{1}{c}{F\ \ } & \multicolumn{1}{c}{P\ \ } & \multicolumn{1}{c|}{R} & \multicolumn{1}{c}{F\ \ } & \multicolumn{1}{c}{P\ \ } & \multicolumn{1}{c|}{R\ \ } & 
  \multicolumn{1}{c}{F\ \ } & \multicolumn{1}{c}{P\ \ } & \multicolumn{1}{c}{R\ \ } \\
  \hline
  MTM-Vote & 63.6 & 64.4 & 62.8 & 61.0 & 56.7 & 65.9 & 50.4 & 44.8 & 57.5 & 83.9 & 88.4 & 79.8 \\
  Unified-01 & 42.7 & 93.7 & 27.6 & 37.5 & 72.5 & 25.3 & 23.6 & 50.8 & 15.4 & 01.6 & 97.8 & 00.8 \\
  Unified-11 & 70.2 & 73.8 & 67.0 & 67.7 & 64.0 & 71.9 & \underline{53.2} & 47.1 & 61.1 & 80.1 & 84.6 & 76.1 \\
  Unified-00 & \textbf{73.8} & 84.1 & 65.7 & \textbf{69.7} & 68.1 & 71.5 & 52.7 & 49.4 & 56.5 & \textbf{84.8} & 90.0 & 80.2 \\
  \bottomrule
\end{tabular}
\caption{Results for task adaptation in the no-supervision setting. The best f1 score in each column that is significantly better than the second best is bold-faced, while those are better but not significantly are underlined. All the significance tests are conducted  using mention-level McNemar's Chi-square test, with p-value $=$ 0.01.} \label{tab:global}
%\vspace{-1em}
\end{table*}

\subsection{Hyper-parameters.}
We borrow most of the best hyper-parameters reported by \newcite{wang2018cross}. The hidden sizes of the BiLSTMs are tuned, and the best value we found is 100 for the character-level BiLSTM, and 300 for the word-level BiLSTM. We also tuned both discounting factors $M$ and $M'$ in the range of [0,0.2,0.4,0.6,0.8,1.0]. It turns out that $M=0, M'=0$ (using improved partition function) and $M=1, M'=1$ (using improved gold energy function) make two local optimums. Therefore we report the performance of three special cases of our proposed framework, with $M, M'=[0,0]$, $[1,1]$, and $[0,1]$ (the naive model), respectively.

\subsection{Compared Models.}
%We benchmark our unified model by comparing it with several baselines. 
We compare different variations of our unified model and other models in different settings.
We first train models on all training corpora, and then perform evaluations under two scenarios: (1) \textbf{no-supervision}: directly evaluating the trained models on each global corpora; (2) \textbf{limited-supervision}: fine-tuning the models on a small subset of the training portion of each global corpus before the evaluations. 
% For the local evaluation, we compare with the state-of-the-art single task NER model \cite{liu2017empower} and multi-task learning methods \cite{crichton2017neural,wang2018cross}.

Under both scenarios, we report performance of four different models:
\begin{itemize}
    \item \textbf{MTM/MTM-vote}: Train a multi-task model (MTM) on training corpora, using a separate CRF for each corpus. (This is the current state-of-the-art structure \cite{wang2018cross} when evaluated on the training corpora.)
    \begin{itemize}
        \item Under the no-supervision setting, we heuristically combine all existing CRF's predictions to make global predictions. Specifically, we apply two heuristics to resolve conflicts while preserving entity chunk-level consistency. First, where predictions from more than one model overlap, we expand each prediction's boundary to the outermost position. Second, we always favor the predictions of named entities over the predictions of non-entity.\footnote{A lower recall and f1 score was observed in the initial experiment without this heuristic.}
        \item Under the limited-supervision setting, for each global corpus, we add a new CRF and train it along with the LSTMs. 
    \end{itemize}
    \item \textbf{Unified-01}: Use the naive training approach described in \ref{sec:naive}; this corresponds to our unified model with settings $M=0,M'=1$.
    \item \textbf{Unified-11}: Use the improved gold energy function described in \ref{sec:improve-energy}; this corresponds to our unified model with settings $M=1,M'=1$ and is equivalent to the model proposed by \newcite{greenberg2018marginal}.
    \item \textbf{Unified-00}: Use the improved partition function proposed in \ref{sec:improve-partition}; this corresponds to our unified model with settings $M=0,M'=0$.
\end{itemize}

Among the compared models, \textit{Unified-01} (the naive model) and \textit{MTM/MTM-Vote} are either simple or commonly used methods and thus are treated as baselines. \textit{Unified-00} is a novel approach. Although \newcite{greenberg2018marginal} used the approach of \textit{Unified-11}, they only evaluated the model on training corpora/tasks while we apply it for task adaptation. Moreover, it is a special case of our proposed framework, thus we argue that people can simply tune $M$ and $M'$ to get good performance for adaptations to new tasks.

\begin{figure*}[t!]
\includegraphics[width=0.5\textwidth]{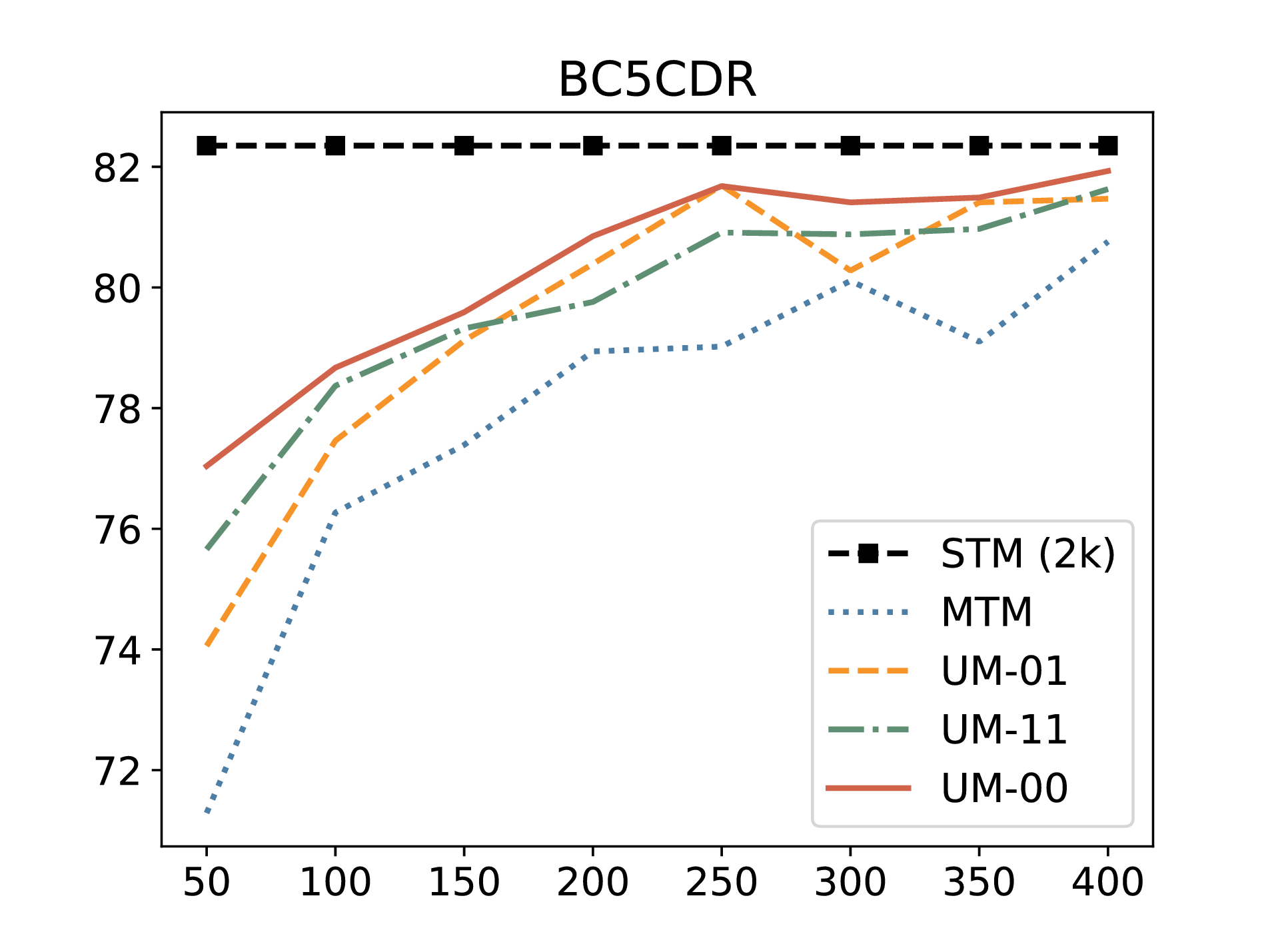}
\includegraphics[width=0.5\textwidth]{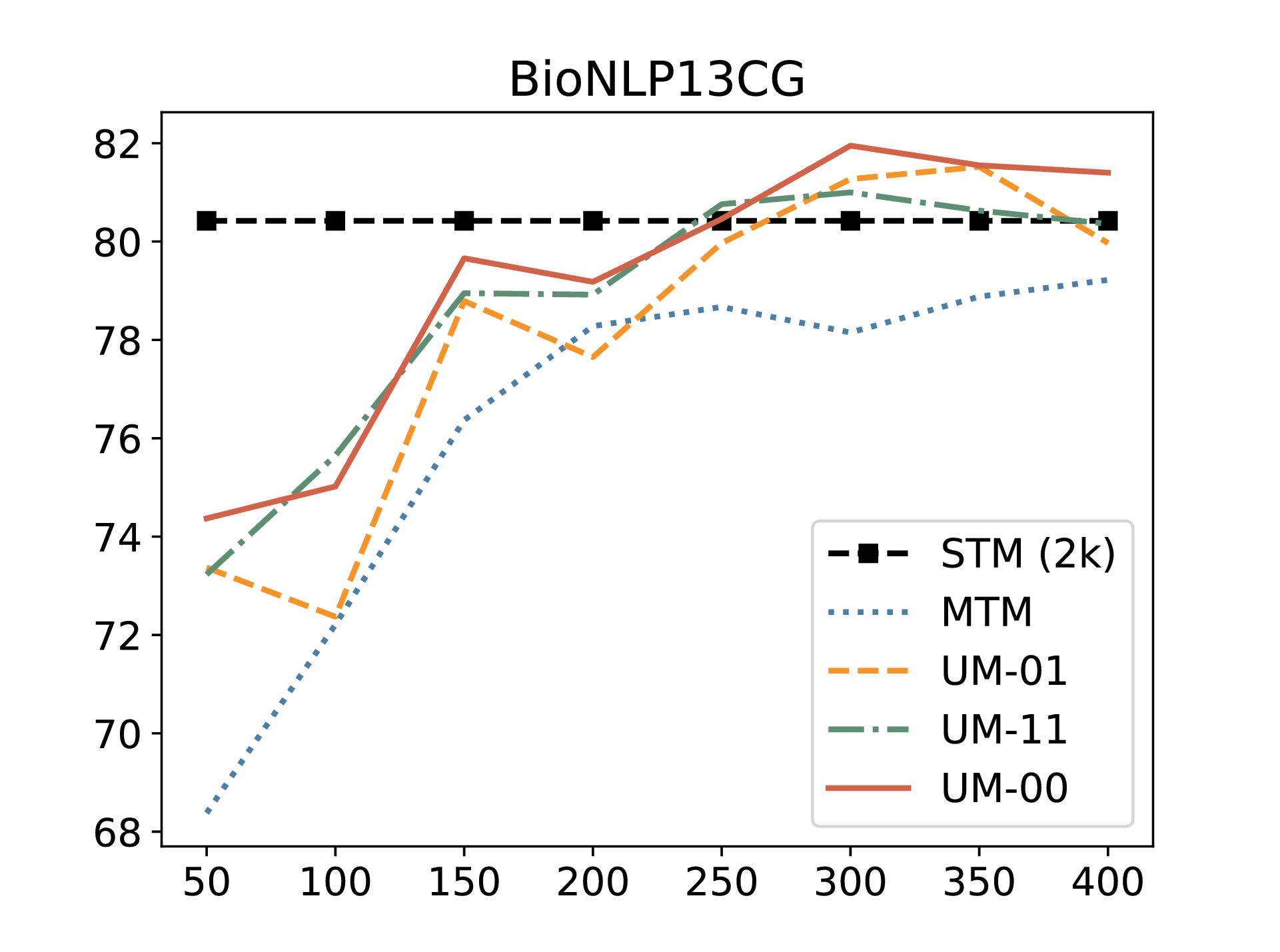}
\includegraphics[width=0.5\textwidth]{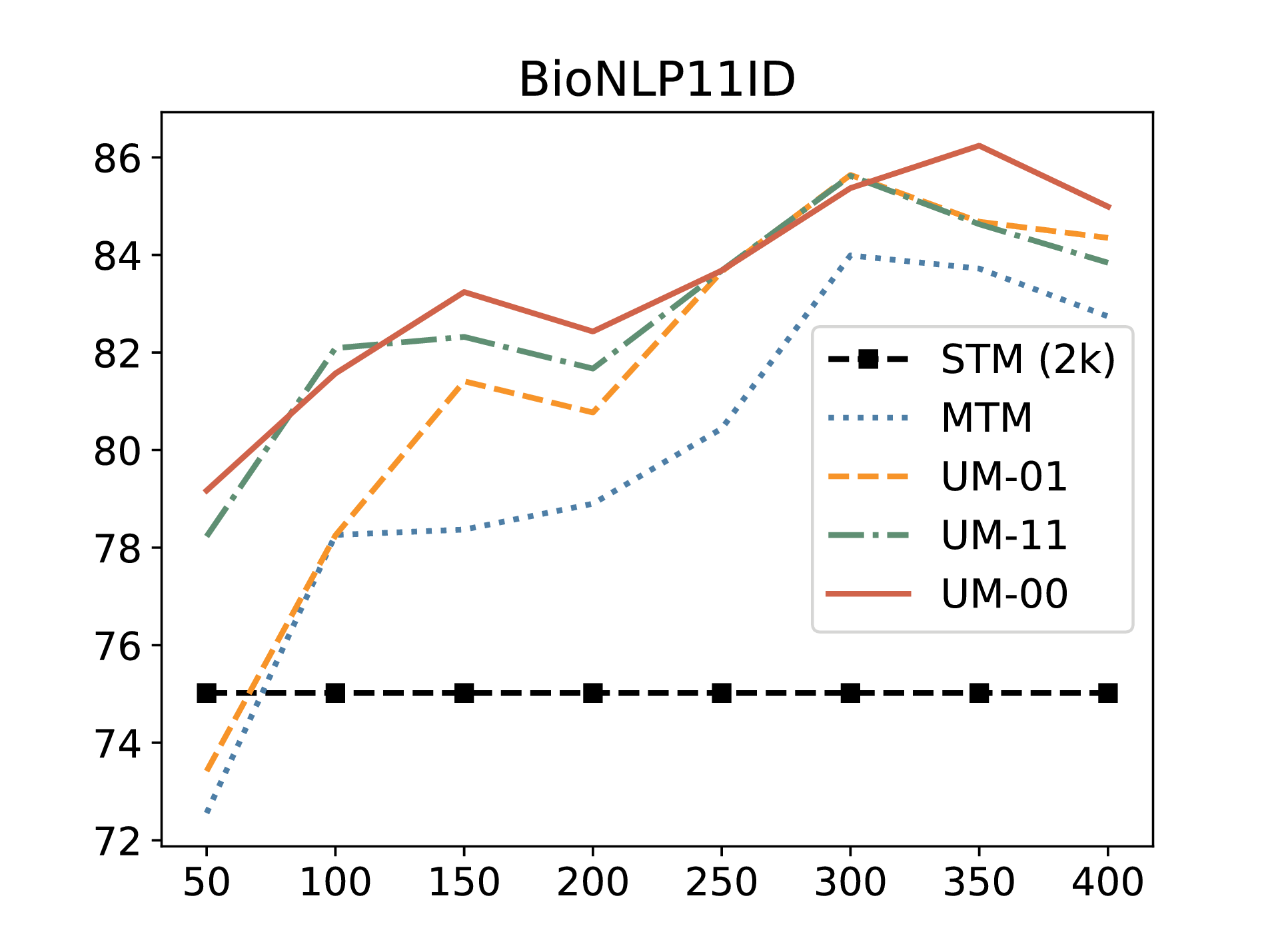}
\includegraphics[width=0.5\textwidth]{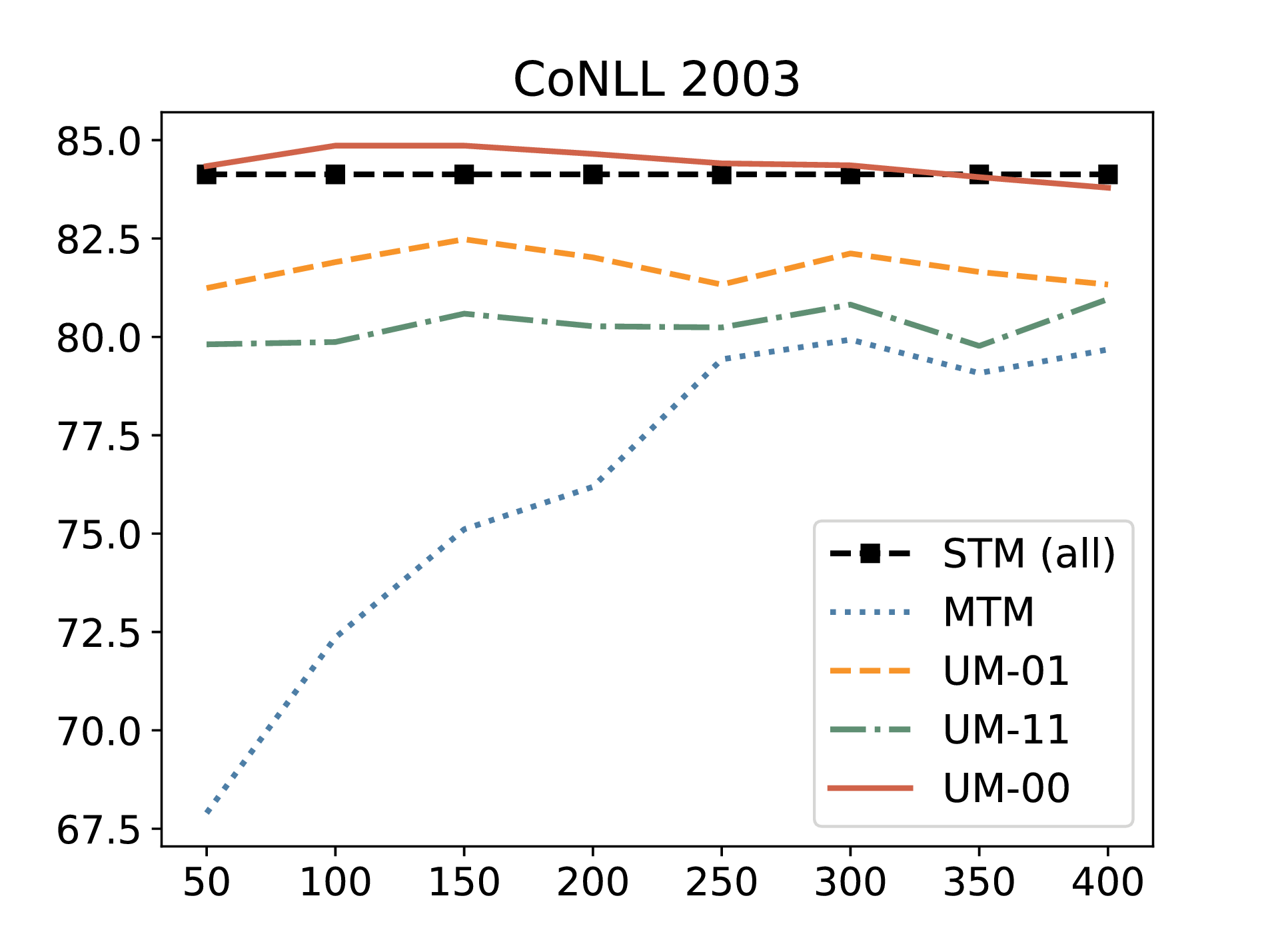}
\caption{Plot of f1 scores for task adaptation in the limited-supervision setting. X-axis represents the number of sentences used for fine-tuning. \textit{STM(2k)} is a STM trained on 2k sentences sampled from the global corpus, and \textit{STM(all)} is trained on the entire training set of the corpus.}
\label{fig:result}
\end{figure*}

\section{Results}
As mentioned above, we compare the results of four different approaches in no-supervision and limited-supervision settings, both with real-world biomedical data and synthetic news data. 

As a sanity check, we also evaluate the models on the test sets of the training corpora. The results can be found in Appendix~\ref{app:local}. It is shown that our MTM performs comparably with state-of-the-art systems evaluated on the training corpora, and thus is a strong baseline.

\subsection{No-Supervision Setting} 
Table~\ref{tab:global} demonstrates the results for task adaptation in the no-supervision setting. We report precision and recall in addition to f1 scores to better show the differences between the models. 

Comparing on f1 scores, \textit{Unified-00} (our new model) significantly outperforms all other models on three out of four datasets, demonstrating its effectiveness. \textit{Unified-11} also achieves good results, with higher recall but lower precision than \textit{Unified-00}. This aligns well with our hypothesis that it encourages predictions of entities. Conversely, \textit{Unified-01} (the naive approach) achieves the highest precision but lowest recall, which is reasonable considering the problem of false penalties that discourages the model from predicting entities. We also found that the model achieves better performance when $M=M'$, which supports our hypothesis in \ref{sec:combine} that the model works better with a valid likelihood function.
%As a reminder, none of our unified models are trained on any dataset that matches the tagset of any test corpora, so the results are not comparable with those reported in e.g.~\newcite{greenberg2018marginal,crichton2017neural,perez2017evaluation}. 

\subsection{Limited-Supervision Setting} 
To further demonstrate the models' ability to adapt to new datasets with a small amount of supervision, we sample a small subset of the training portion of each global evaluation corpus to fine-tune the trained models. We show the performance of the models fine-tuned with different amounts of sampled data. For each global corpus, we show a single-task model (STM) trained on it with a reasonable amount of data (two thousand sentences for the biomedical corpora). In the CoNLL 2003 setting, we train the STM on the entire training data for a fair comparison, because all other models are first trained on the four training portions, which essentially look through the entire training set (just partially annotated). The results of the STMs are used as benchmarks. Experimental results are presented in Figure~\ref{fig:result}.

Firstly, with much less training data, all the models achieve comparable or noticeably better performance than the STMs trained from scratch, demonstrating that training on the partially annotated corpora does help to boost performance on global evaluation corpora. Additionally, MTMs are worse than all the unified models, because they only share the LSTM layers, but lose all the knowledge in the CRFs when adapted to new corpora. The unified models have the advantage that they can reuse the robust CRFs learned from a large amount of data. This is more obvious in the CoNLL 2003 evaluation setting, where the unified models that reuse the pre-trained CRFs achieve good performance trained with only 50 sentences, but the MTM, which does not reuse the CRFs, needs a larger amount of training data to catch up. 

In general, \textit{Unified-00}, our novel approach proposed here, still performs the best on every dataset. We note that although \textit{Unified-01} has an extremely low recall on the CoNLL 2003 dataset in the no-supervision setting, it works surprisingly well in the limited-supervision setting. On the other hand, \textit{Unified-00} and \textit{Unified-11} generally perform better than \textit{Unified-01} on real-world biomedical datasets, especially when fine-tuned on less data. Again, since all the unified models are special cases of our proposed framework, we argue that, for adapting to new datasets, people can simply tune the discounting factors $M$ and $M'$ to get good results. 

\section{Conclusion and Future Work}
In this paper, we propose a unified model that learns from multiple partially annotated datasets to make {\em joint} predictions on the union of entity types appearing in any training dataset.
The model integrates learning signals from different datasets and avoids potential conflicts that would result from combining independent predictions from multiple models. 
Experiments show that the proposed unified model can efficiently adapt to new corpora that have more entity types than any of the training corpora, and performs better than the baseline approaches.

In future work, we plan to explore other algorithms (e.g.\ imitation learning) that allow the model the explore the unknown space during training, using delayed rewards to decide whether the model should trust its exploration. 
Analysis of the global evaluation results suggests that the unified model is under-predicting, meaning there is still room for improvement specifically on recall. We plan to explore further changes to the current objectives to encourage more entity predictions.

Finally, the approach proposed in this paper also does not handle entity types of varying granularities or tagsets with mismatched guidelines (e.g.\ one dataset annotates only for-profit companies as \textit{ORG} and one annotates all formalized groups). Effectively modeling these complications is an interesting area for future work. 

\section*{Acknowledgements}
We thank the anonymous reviewers for their constructive comments, as well as the members of
the USC PLUS lab for their early feedbacks. We thank Tianyu Meng and Yuxin Zhou for their help with initial data processing and experimental setup. 
This work is supported in part by DARPA
(HR0011-15-C-0115) and an NIH R01 (LM012592). Approved for Public Release, Distribution Unlimited. The views expressed are those of the authors and do not reflect the official policy or position of the sponsors.

\bibliographystyle{acl_natbib}
\bibliography{emnlp-ijcnlp-2019,appendix_ref}

\begin{thebibliography}{44}
\expandafter\ifx\csname natexlab\endcsname\relax\def\natexlab#1{#1}\fi

\bibitem[{Aguilar et~al.(2014)Aguilar, Beller, McNamee, V.~Durme, Strassel,
  Song, and Ellis}]{aguilar2014comparison}
Jacqueline Aguilar, Charley Beller, Paul McNamee, Ben V.~Durme, Stephanie
  Strassel, Zhiyi Song, and Joe Ellis. 2014.
\newblock A comparison of the events and relations across {ACE}, {ERE},
  {TAC-KBP}, and {FrameNet} annotation standards.
\newblock In \emph{ACL Workshop: EVENTS}.

\bibitem[{Bellare and McCallum(2007)}]{bellare2007learning}
Kedar Bellare and Andrew McCallum. 2007.
\newblock Learning extractors from unlabeled text using relevant databases.
\newblock In \emph{Sixth international workshop on information integration on
  the web}.

\bibitem[{Bengio et~al.(1994)Bengio, Simard, and Frasconi}]{bengio1994learning}
Yoshua Bengio, Patrice Simard, and Paolo Frasconi. 1994.
\newblock Learning long-term dependencies with gradient descent is difficult.
\newblock \emph{IEEE Transactions on Neural Networks}.

\bibitem[{Carlson et~al.(2009)Carlson, Gaffney, and
  Vasile}]{Carlson2009LearningAN}
Andrew Carlson, Scott Gaffney, and Flavian Vasile. 2009.
\newblock Learning a named entity tagger from gazetteers with the partial
  perceptron.
\newblock In \emph{AAAI Spring Symposium: Learning by Reading and Learning to
  Read}.

\bibitem[{Consortium(2013)}]{ere13}
Linguistic~Data Consortium. 2013.
\newblock {DEFT ERE} annotation guidelines: Relations v1.1.
\newblock \emph{Linguistic Data Consortium, Philadelphia}.

\bibitem[{Craven and Kumlien(1999)}]{craven1999constructing}
Mark Craven and Johan Kumlien. 1999.
\newblock Constructing biological knowledge bases by extracting information
  from text sources.
\newblock In \emph{ISMB}.

\bibitem[{Craven et~al.(1998)Craven, McCallum, PiPasquo, Mitchell, and
  Freitag}]{craven1998learning}
Mark Craven, Andrew McCallum, Dan PiPasquo, Tom Mitchell, and Dayne Freitag.
  1998.
\newblock Learning to extract symbolic knowledge from the world wide web.
\newblock Technical report, Carnegie-mellon univ pittsburgh pa school of
  computer Science.

\bibitem[{Crichton et~al.(2017)Crichton, Pyysalo, Chiu, and
  Korhonen}]{crichton2017neural}
Gamal Crichton, Sampo Pyysalo, Billy Chiu, and Anna Korhonen. 2017.
\newblock A neural network multi-task learning approach to biomedical named
  entity recognition.
\newblock \emph{BMC bioinformatics}.

\bibitem[{Do{\u{g}}an et~al.(2014)Do{\u{g}}an, Leaman, and Lu}]{dougan2014ncbi}
Rezarta~Islamaj Do{\u{g}}an, Robert Leaman, and Zhiyong Lu. 2014.
\newblock Ncbi disease corpus: a resource for disease name recognition and
  concept normalization.
\newblock \emph{Journal of biomedical informatics}.

\bibitem[{Fernandes and Brefeld(2011)}]{fernandes2011learning}
Eraldo~R Fernandes and Ulf Brefeld. 2011.
\newblock Learning from partially annotated sequences.
\newblock In \emph{Joint European Conference on Machine Learning and Knowledge
  Discovery in Databases}, pages 407--422. Springer.

\bibitem[{Gerner et~al.(2010)Gerner, Nenadic, and Bergman}]{gerner2010linnaeus}
Martin Gerner, Goran Nenadic, and Casey~M Bergman. 2010.
\newblock Linnaeus: a species name identification system for biomedical
  literature.
\newblock \emph{BMC bioinformatics}.

\bibitem[{Greenberg et~al.(2018)Greenberg, Bansal, Verga, and
  McCallum}]{greenberg2018marginal}
Nathan Greenberg, Trapit Bansal, Patrick Verga, and Andrew McCallum. 2018.
\newblock Marginal likelihood training of bilstm-crf for biomedical named
  entity recognition from disjoint label sets.
\newblock In \emph{Proceedings of the 2018 Conference on Empirical Methods in
  Natural Language Processing}, pages 2824--2829.

\bibitem[{Hasegawa et~al.(2004)Hasegawa, Seki, and
  Grishman}]{hasegawa2004discovering}
Takaaki Hasegawa, Satoshi Seki, and Ralph Grishman. 2004.
\newblock Discovering relations among named entities from large corpora.
\newblock In \emph{ACL}.

\bibitem[{Hochreiter and Schmidhuber(1997)}]{hochreiter1997long}
Sepp Hochreiter and J{\"u}rgen Schmidhuber. 1997.
\newblock Long short-term memory.
\newblock \emph{Neural computation}.

\bibitem[{Ji et~al.(2010)Ji, Grishman, Dang, Griffitt, and
  Ellis}]{ji2010overview}
Heng Ji, Ralph Grishman, Hoa~Trang Dang, Kira Griffitt, and Joe Ellis. 2010.
\newblock Overview of the tac 2010 knowledge base population track.
\newblock In \emph{TAC}.

\bibitem[{Kim et~al.(2004)Kim, Ohta, Tsuruoka, Tateisi, and
  Collier}]{kim2004introduction}
Jin-Dong Kim, Tomoko Ohta, Yoshimasa Tsuruoka, Yuka Tateisi, and Nigel Collier.
  2004.
\newblock Introduction to the bio-entity recognition task at jnlpba.
\newblock In \emph{JNLPBA}. ACL.

\bibitem[{Kim et~al.(2013)Kim, Wang, and Yasunori}]{kim2013genia}
Jin-Dong Kim, Yue Wang, and Yamamoto Yasunori. 2013.
\newblock The genia event extraction shared task, 2013 edition-overview.
\newblock In \emph{BioNLP Shared Task 2013 Workshop}.

\bibitem[{Krallinger et~al.(2015)Krallinger, Leitner, Rabal, Vazquez,
  Oyarzabal, and Valencia}]{krallinger2015chemdner}
Martin Krallinger, Florian Leitner, Obdulia Rabal, Miguel Vazquez, Julen
  Oyarzabal, and Alfonso Valencia. 2015.
\newblock Chemdner: The drugs and chemical names extraction challenge.
\newblock \emph{Journal of Cheminfo.}

\bibitem[{Lafferty et~al.(2001)Lafferty, McCallum, and
  Pereira}]{lafferty2001conditional}
John Lafferty, Andrew McCallum, and Fernando~CN Pereira. 2001.
\newblock Conditional random fields: Probabilistic models for segmenting and
  labeling sequence data.
\newblock In \emph{ICML}.

\bibitem[{Lample et~al.(2016)Lample, Ballesteros, Subramanian, Kawakami, and
  Dyer}]{lample2016neural}
Guillaume Lample, Miguel Ballesteros, Sandeep Subramanian, Kazuya Kawakami, and
  Chris Dyer. 2016.
\newblock Neural architectures for named entity recognition.
\newblock In \emph{NAACL}.

\bibitem[{Li and Liu(2005)}]{li2005learning}
Xiao-Li Li and Bing Liu. 2005.
\newblock Learning from positive and unlabeled examples with different data
  distributions.
\newblock In \emph{ECML}.

\bibitem[{{Liu} et~al.(2018){Liu}, {Shang}, {Xu}, {Ren}, {Gui}, {Peng}, and
  {Han}}]{liu2017empower}
L.~{Liu}, J.~{Shang}, F.~{Xu}, X.~{Ren}, H.~{Gui}, J.~{Peng}, and J.~{Han}.
  2018.
\newblock {Empower Sequence Labeling with Task-Aware Neural Language Model}.
\newblock In \emph{AAAI}.

\bibitem[{Liu et~al.(2014)Liu, Zhang, Che, Liu, and Wu}]{D14-1093}
Yijia Liu, Yue Zhang, Wanxiang Che, Ting Liu, and Fan Wu. 2014.
\newblock \href {https://doi.org/10.3115/v1/D14-1093} {Domain adaptation for
  crf-based chinese word segmentation using free annotations}.
\newblock In \emph{Proceedings of the 2014 Conference on Empirical Methods in
  Natural Language Processing (EMNLP)}, pages 864--874. Association for
  Computational Linguistics.

\bibitem[{Lu et~al.(2015)Lu, Ji, Yao, Wei, and Liang}]{lu2015chemdner}
Yanan Lu, Donghong Ji, Xiaoyuan Yao, Xiaomei Wei, and Xiaohui Liang. 2015.
\newblock Chemdner system with mixed conditional random fields and multi-scale
  word clustering.
\newblock \emph{Journal of cheminformatics}.

\bibitem[{Ma and Hovy(2016)}]{ma2016end}
Xuezhe Ma and Eduard Hovy. 2016.
\newblock End-to-end sequence labeling via bi-directional lstm-cnns-crf.
\newblock In \emph{ACL}.

\bibitem[{McCallum and Li(2003)}]{mccallum2003early}
Andrew McCallum and Wei Li. 2003.
\newblock Early results for named entity recognition with conditional random
  fields, feature induction and web-enhanced lexicons.
\newblock In \emph{NAACL}.

\bibitem[{Mooney and Bunescu(2005)}]{mooney2005subsequence}
Raymond~J Mooney and Razvan~C Bunescu. 2005.
\newblock Subsequence kernels for relation extraction.
\newblock In \emph{NIPS}.

\bibitem[{Neves et~al.(2012)Neves, Damas, Kurtz, and
  Leser}]{neves2012annotating}
Mariana Neves, Alexander Damas, Andreas Kurtz, and Ulf Leser. 2012.
\newblock Annotating and evaluating text for stem cell research.
\newblock In \emph{BioTxtM workshop at LREC on Building and Evaluation
  Resources.}

\bibitem[{Peng and Dredze(2015)}]{pengnamed}
Nanyun Peng and Mark Dredze. 2015.
\newblock Named entity recognition for chinese social media with jointly
  trained embeddings.
\newblock In \emph{Proceedings of the 2015 Conference on Empirical Methods in
  Natural Language Processing (EMNLP)}, Lisboa, Portugal.

\bibitem[{Peng and Dredze(2016)}]{pengimproving}
Nanyun Peng and Mark Dredze. 2016.
\newblock Improving named entity recognition for chinese social media via
  learning segmentation representations.
\newblock In \emph{ACL}.

\bibitem[{Peng and Dredze(2017)}]{peng2016multi}
Nanyun Peng and Mark Dredze. 2017.
\newblock Multi-task domain adaptation for sequence tagging.
\newblock In \emph{Proceddings of the ACL Workshop on Representation Learning
  for NLP}.

\bibitem[{Pyysalo and Ananiadou(2013)}]{pyysalo2013anatomical}
Sampo Pyysalo and Sophia Ananiadou. 2013.
\newblock Anatomical entity mention recognition at literature scale.
\newblock \emph{Bioinformatics}.

\bibitem[{Ritter et~al.(2011)Ritter, Clark, Etzioni et~al.}]{ritter2011named}
Alan Ritter, Sam Clark, Oren Etzioni, et~al. 2011.
\newblock Named entity recognition in tweets: an experimental study.
\newblock In \emph{EMNLP}.

\bibitem[{Sang and De~Meulder(2003)}]{sang2003introduction}
Erik~F Sang and Fien De~Meulder. 2003.
\newblock Introduction to the conll-2003 shared task: Language-independent
  named entity recognition.
\newblock \emph{arXiv preprint cs/0306050}.

\bibitem[{Shang et~al.(2018)Shang, Liu, Ren, Gu, Ren, and
  Han}]{shang2018learning}
Jingbo Shang, Liyuan Liu, Xiang Ren, Xiaotao Gu, Teng Ren, and Jiawei Han.
  2018.
\newblock Learning named entity tagger using domain-specific dictionary.
\newblock \emph{arXiv preprint arXiv:1809.03599}.

\bibitem[{Smith et~al.(2008)Smith, Tanabe, nee Ando, Kuo, Chung, Hsu, Lin,
  Klinger, Friedrich, Ganchev et~al.}]{smith2008overview}
Larry Smith, Lorraine~K Tanabe, Rie~Johnson nee Ando, Cheng-Ju Kuo, I-Fang
  Chung, Chun-Nan Hsu, Yu-Shi Lin, Roman Klinger, Christoph~M Friedrich, Kuzman
  Ganchev, et~al. 2008.
\newblock Overview of biocreative ii gene mention recognition.
\newblock \emph{Genome biology}.

\bibitem[{Soon et~al.(2001)Soon, Ng, and Lim}]{soon2001machine}
Wee~Meng Soon, Hwee~Tou Ng, and Daniel Chung~Yong Lim. 2001.
\newblock A machine learning approach to coreference resolution of noun
  phrases.
\newblock \emph{Computational linguistics}.

\bibitem[{Tjong Kim~Sang and De~Meulder(2003)}]{tjong2003introduction}
Erik~F Tjong Kim~Sang and Fien De~Meulder. 2003.
\newblock Introduction to the conll-2003 shared task: Language-independent
  named entity recognition.
\newblock In \emph{CoNLL at HLT-NAACL}.

\bibitem[{Tsuboi et~al.(2008)Tsuboi, Kashima, Mori, Oda, and
  Matsumoto}]{C08-1113}
Yuta Tsuboi, Hisashi Kashima, Shinsuke Mori, Hiroki Oda, and Yuji Matsumoto.
  2008.
\newblock \href {http://aclweb.org/anthology/C08-1113} {Training conditional
  random fields using incomplete annotations}.
\newblock In \emph{Proceedings of the 22nd International Conference on
  Computational Linguistics (Coling 2008)}, pages 897--904. Coling 2008
  Organizing Committee.

\bibitem[{Walker et~al.(2006)Walker, Strassel, Medero, and
  Maeda}]{walker2006ace}
Christopher Walker, Stephanie Strassel, Julie Medero, and Kazuaki Maeda. 2006.
\newblock Ace 2005 multilingual training corpus.
\newblock \emph{LDC}.

\bibitem[{Wang et~al.(2018)Wang, Zhang, Ren, Zhang, Zitnik, Shang, Langlotz,
  and Han}]{wang2018cross}
Xuan Wang, Yu~Zhang, Xiang Ren, Yuhao Zhang, Marinka Zitnik, Jingbo Shang,
  Curtis Langlotz, and Jiawei Han. 2018.
\newblock Cross-type biomedical named entity recognition with deep multi-task
  learning.
\newblock \emph{CoRR}.

\bibitem[{Wei et~al.(2015)Wei, Peng, Leaman, Davis, Mattingly, Li, Wiegers, and
  Lu}]{wei2015overview}
Chih-Hsuan Wei, Yifan Peng, Robert Leaman, Allan~Peter Davis, Carolyn~J
  Mattingly, Jiao Li, Thomas~C Wiegers, and Zhiyong Lu. 2015.
\newblock Overview of the biocreative v chemical disease relation (cdr) task.
\newblock In \emph{BC V Workshop}.

\bibitem[{Yang and Vozila(2014)}]{Yang2014SemiSupervisedCW}
Fan Yang and Paul Vozila. 2014.
\newblock Semi-supervised chinese word segmentation using partial-label
  learning with conditional random fields.
\newblock In \emph{EMNLP}.

\bibitem[{Yang et~al.(2018)Yang, Chen, Li, He, and Zhang}]{yang2018distantly}
Yaosheng Yang, Wenliang Chen, Zhenghua Li, Zhengqiu He, and Min Zhang. 2018.
\newblock Distantly supervised ner with partial annotation learning and
  reinforcement learning.
\newblock In \emph{Proceedings of the 27th International Conference on
  Computational Linguistics}, pages 2159--2169.

\end{thebibliography}

\clearpage
\appendix
\section{Appendix} 

\subsection{Datasets} \label{app:data}

\begin{table}[t]
\centering
\small
\begin{tabular}{l@{\hspace{\tabcolsep}}m{2.2cm}@{ }l@{\hspace{\tabcolsep}}l@{\hspace{\tabcolsep}}l} 
\toprule
Corpus &   Named Entities  & Sents & Tokens & Mentions \\ 
\midrule
BC2GM & Gene/Protein & 20,131 & 569,912 & 24,585 \\ \midrule
BC4CHEM & Chemical & 87,685 & 2,544,305 &  84,312\\ \midrule
NCBI & Disease & 7,287 & 184,167 &  6,883\\ \midrule
JNLPBA & \makecell[l]{Gene/Protein,\\ DNA, \\Cell-type, \\Cell-line, \\RNA} & 24,806 & 595,994 & 59,965 \\ \midrule
%AnatEM & Anatomy & 11809 & 312584 &  13703 \\ \midrule
Linnaeus & Species & 23,155 & 539,428 & 4,265 \\ 
\bottomrule
\end{tabular}
\caption{Statistics for the Training Corpora} \label{tab:train-data}
\vspace{-1.5em}
\end{table}

\begin{table}[t]
\centering
\small
\begin{tabular}{p{1.7cm}m{2cm}@{\hspace{\tabcolsep}}l@{\hspace{\tabcolsep}}l@{\hspace{\tabcolsep}}l}
  \toprule
  Corpus &  Named Entities & Sents & Tokens & Mentions\\
  \hline
  BC5CDR &  \makecell[l]{Chemical, \\Disease}  & 13,938 & 360,373 &  28,789 \\ \hline
  BioNLP13CG & \makecell[l]{Gene/Protein,  \\Disease, \\Chemical, \\Others} & 1,906 & 52,771 & 6881 \\ \hline
  BioNLP11ID & \makecell[l]{Gene/Protein, \\Chemical, \\Others} & 5178 & 166416 & 11084 \\
  \bottomrule
\end{tabular}
\footnotetext{Footnote}
\caption{Statistics for global evaluation corpora. ``Others'' denote the NEs which do not appeared in training data, thus are not evaluated.} \label{tab:test-data}
\vspace{-1em}
\end{table}

Below we introduce the datasets in the biomedicine domain and the news domain.

\subsubsection{Biomedicine domain: Local training group} \label{app:train}
The training group consists of five datasets: \textit{BC2GM}, \textit{BC4CHEM}, \textit{NCBI-disease}, \textit{JNLPBA}, and \textit{Linnaeus}. The first two datasets 
are from different BioCreative shared tasks \cite{smith2008overview,krallinger2015chemdner,wei2015overview}. \textit{NCBI-disease} is created by \newcite{dougan2014ncbi} for disease name recognition and normalization. 
\textit{JNLPBA} comes from the 2004 shared task from joint workshop on natural language processing in biomedicine and its applications \cite{kim2004introduction}, and \textit{Linnaeus} is a species corpus composed by \newcite{gerner2010linnaeus}. More information about the datasets can be found in Table~\ref{tab:train-data}. 

Below are detailed descriptions of the datasets:

\textbf{BC2GM} is a gene/protein corpus. The annotation is Gene. It's provided by the BioCreative II Shared Task for gene mention recognition. 
%The state-of-art system is proposed by \cite{ando2007biocreative}. The original shared task using an alternative match criterion, achieving a 87.21\% F1 score. In our experiment we compare with \cite{crichton2017neural} \newcite{wang2018cross} F1 score based on exact match.

\textbf{BC4CHEM} is a chemical corpus. The annotation is Chemical. It's provided by the BioCreative IV Shared Task for chemical mention recognition. 
%The state of art benchmark system is proposed by \cite{lu2015chemdner}, achieving a 88.06\% F1 score.

\textbf{NCBI-disease} is a disease corpus. The annotation is Disease. It was introduced for disease name recognition and normalization. 
%The state of art benchmark system is TaggerOne \cite{leaman2016taggerone} with 82.90 F1 score.

\textbf{JNLPBA} consists of DNA, RNA, Gene/Protein, Cell line, Cell Type. The annotation is same as the NE names, except the Gene/Protein is annotated with Protein. It was provided by 2004 JNLPBA Shared Task for biomedical entity recognition. 
%The state-of-the-art system is \cite{Zhou04} with a 72.55\% F1 score. Although this system is a little out-of-date, it still serves as a decent benchmark.

\textbf{Linnaeus} is a species corpus. The annotation is Species. The original project was created for entity mention recognition. %and released with NER system with 95.68\% F1 score. However the system is a dictionary based method and the performance was evaluated on the entire corpus instead of test set alone. Also a later work reported\cite{pafilis2013species} a significant lower F1 score(85.1\%) for this system while their model achieves 91.1\%. Thus we consider 91.1\% as the benchmark. 

\begin{table}[t]
\centering
\small
\begin{tabular}{l l l l}
  \toprule
  & Articles &  Sentences & Tokens \\
  \hline
  Training set & 946 & 14,987 & 203.621 \\ \midrule
  Development set & 216 & 3,466 & 51,362 \\ \midrule
  Test set & 231 & 3,684 & 46,435 \\ 
  \bottomrule
\end{tabular}
\footnotetext{Footnote}
\caption{Statistics for the CoNLL 2003 NER dataset} \label{tab:Conll2003_data}
\end{table}

\begin{table*}[ht!]
\centering
%\resizebox{\linewidth}{!}{
\begin{tabular}{@{ }l@{ }|c@{\ \ }c@{\ \ }c@{\ \ }c@{\ \ }c@{\ \ }}
  \toprule
  %\multirow{2}{*}{\textbf{Corpus}} & \multicolumn{7}{c}{\bf Trained on Other Biomedical Datasets} \\ \cline{2-8}
  \textbf{Corpus} & \bf BC2GM & \bf BC4CHM & \bf NCBI & \bf JNLPBA & \bf Linnaeus \\
  \hline
  STM & 79.9 & 88.6 & 84.1 & 72.7 & 87.3 \\
  MTM \protect\newcite{crichton2017neural} & 73.2 & 83.0 & 80.4 & 70.1 & 84.0 \\
  MTM \protect\newcite{wang2018cross} & \underline{80.7} & \underline{89.4} & \underline{86.1} & 73.5 & -  \\ 
  MTM (ours) & 80.3 & 89.2 & 85.8 & 73.5 & \textbf{88.5}  \\
  Unified-01 & 70.9 & 83.5 & 79.8 & 80.9 & 79.9 \\
  Unified-11 & 74.2 & 84.1 & 80.5 & 80.9 & 80.7 \\
  \hline
  Unified-00 & 79.1 & 87.3 & 84.0 & \textbf{83.8} & 83.9 \\
  \bottomrule
\end{tabular}%}
\caption[]{Local evaluation (f1 scores). The best results that are significantly better than the second best are bold-faced, while those are best but not significantly better than the second best are underlined. All the significance tests are conducted  using mention-level McNemar's Chi-square test, with p-value $=$ 0.01.
} \label{tab:local}
\end{table*}

\subsubsection{Biomedicine domain: Global evaluation group} \label{app:test}
We reemphasize here that the purpose of the global evaluation is to test the model's ability to making global predictions and efficiently adapt to global corpora. 
While no corpus is globally annotated, we identify several existing corpora to {\em approximate} the global evaluation. Each test corpus is annotated with a {\em superset} of several training corpora to test the model's generalizability outside of the local tag spaces. 

The global evaluation group contains three datasets: {\em BC5CDR}, {\em BioNLP13CG}, and {\em BioNLP11ID}. Each is annotated with multiple entity types. {\em BC5CDR} comes from the BioCreative shared tasks \cite{smith2008overview,krallinger2015chemdner,wei2015overview}. {\em BioNLP13CG} and {\em BioNLP11ID} come from the BioNLP shared task \cite{kim2013genia}. More information about the global evaluation datasets can be found in Table~\ref{tab:test-data}. 

Below are detailed descriptions of the datasets:

\textbf{BC5CDR} is a chemical and disease corpus. The annotation is Chemical and Disease. It's provided by BioCreative V Shared Task for chemical and disease mention recognition. 
%The state of the art system is proposed by \cite{leaman2016taggerone}, achieving a 86.76\% F1 score.

\textbf{BioNLP13CG} consists of Gene/Protein and Related Product, Cancel, Chemical, Anatomy and Organism and others. \textbf{BioNLP11ID} consists of Gene/Protein, Chemical, and Organism. The annotation is same as the NE types but has a finer ontology scope. 
%BioNLP13CG task focuses on cancer, emphasizing the relation extraction of physiological and pathological processes at various levels of biological organization. 

There are inconsistencies between the entity type names in different datasets, mainly due to different granularities. To remove this unnecessary noise, we manually merged some entity types. For example, we unify Gene and Protein into Gene/Protein as they are commonly used interchangeably; we merge ``Simple Chemical'' to ``Chemical'' and leave the problem of entity type granularity for future work. The information in Table~\ref{tab:train-data} and~\ref{tab:test-data} reflects the merged types.

\subsubsection{News domain: CoNLL 2003 NER dataset}
We use the CoNLL 2003 NER dataset (\cite{sang2003introduction}) to evaluate the models in news domain. More information about the dataset can be found in Table \ref{tab:Conll2003_data}. We use synthetic data from the dataset to simulate local training and global evaluation. Specifically, the CoNLL 2003 NER dataset is annotated with four entity types: location, person, organization, and miscellaneous entities. We randomly split the training set into four portions, each contains only one entity type respectively, with other types changed to "O". The models are trained on the four training portions and we test on the original test set with all entity types annotated.

\subsubsection{Data split} 
For the news domain, we use the default train, dev, test portion of the CoNLL 2003 NER dataset. For the biomedicine domain, we follow the data split in \newcite{crichton2017neural} for both the training and the evaluation groups. All datasets are divided into three portions: train, dev, and test. We train the model on the training set of the training group and tune the hyper-parameters on the corresponding development set. Global evaluations are performed on the test set of the evaluation group.

\subsection{Local Evaluation} \label{app:local}

For a sanity check, we evaluate the models on the training corpora and compare the results with state-of-the-art systems. In this setting, all the models are trained on the training set of the training corpora (without fine-tuning on global evaluation corpora) and evaluated on their test set. The results are shown in Table~\ref{tab:local}. \textbf{STM} is the single-task models we implemented, following the settings in \newcite{wang2018cross}. The SOTA is achieved by \newcite{wang2018cross} with multi-task model, which is shown in the table as \textbf{MTM \newcite{wang2018cross}}. They trained their model on \textit{BC2GC}, \textit{BC4CHM}, \textit{NCBI}, \textit{JNLPBA}, and \textit{BC5CDR}. \textbf{MTM (ours)} is the multi-task model we trained on our five training corpora and used as a baseline in the global evaluations. It has the same architecture as \newcite{wang2018cross}.

As we can see, \textbf{MTM \newcite{wang2018cross}} achieves the best results on 3 out of 4 datasets. And our MTM achieves very similar results, showing it is a strong model on training corpora. Our proposed models do not perform very well when evaluated on the training corpora. But in the global evaluation setting, they perform much better compared to our strong MTM. This demonstrates the superiority of our proposed models on task adaptation.

\end{document}